\documentclass[acmtog]{acmart}
\AtBeginDocument{%
  \providecommand\BibTeX{{%
    \normalfont B\kern-0.5em{\scshape i\kern-0.25em b}\kern-0.8em\TeX}}}

\setcopyright{acmcopyright}
\copyrightyear{2023}
\acmYear{2023}
\acmDOI{10.1145/1122445.1122455}

\acmJournal{TOG}
\acmVolume{37}
\acmNumber{4}
\acmArticle{111}
\acmMonth{3}
\usepackage{multirow}
\usepackage{subfigure}
\usepackage{booktabs}
\usepackage{tabularx}
\usepackage{adjustbox}
\usepackage{lipsum} 
\usepackage{tcolorbox}
\usepackage{wrapfig}
\usepackage{diagbox}
\usepackage{fancyhdr}
\usepackage{graphicx}
\newcommand{\ci}[1]{{\color{blue} #1}}

\begin{document}

\title{GPT-Neo for commonsense reasoning - a theoretical and practical lens}

\author{Rohan Kashyap}
\authornote{Both authors contributed equally to this research.}
\orcid{9632145522}
\author{Vivek Kashyap}
\authornotemark[1]
\author{Narendra C.P}
\email{narendracp@bit-bangalore.edu.in}
\affiliation{%
  \institution{Bangalore Institute of Technology}
  \city{Bangalore}
  \country{India}
}


\renewcommand{\shortauthors}{GPT-Neo for commonsense reasoning - a theoretical and practical lens}

\begin{abstract}
Recent work has demonstrated substantial gains in pre-training large-language models (LLMs) followed by supervised fine-tuning on the downstream task. In this paper, we evaluate the performance of the GPT-neo model using $6$ commonsense reasoning benchmark tasks. We aim to examine the performance of smaller models using the GPT-neo models against several larger model baselines such as GPT-$3$, Llama-$2$, MPT and Falcon. Upon fine-tuning with the appropriate set of hyperparameters, our model achieves competitive accuracy on several tasks. We also investigate and substantiate our results using attention-head visualization to better understand the model performance. Finally, we conduct various robustness tests using various methods to gauge the model performance under numerous settings.
\end{abstract}

\ccsdesc[500]{Neural Networks~Large Language models}
\ccsdesc{Reasoning}

\keywords{Transformers}

\maketitle

\section{Introduction}
Commonsense reasoning encompasses a central role in neural language understanding. Recent years has seen the rise of large language models (LLMs), such as the GPT-$3$ \cite{b10}, T5 \cite{b30}, Falcon \cite{b47} and Llama \cite{b25, b26}. Autoregressive language models pretrained on large corpus of self-supervised data using Reinforcement Learning with Human Feedback (RHLF) results in model alignment with respect to human preferences on numerous tasks. Advances in natural language processing tasks such as reading comprehension, question answering and inductive reasoning demonstrate the effectiveness of these huge pre-trained models on numerous downstream tasks, and methods such as instructive fine-tuning \cite{b16} and chain-of-thought prompting \cite{b15} have shown excellent few-shot learning capabilities.\\ 

\noindent In particular, \cite{b15} demonstrates the ability of LLMs for multiple-step reasoning on complex arithmetic and symbolic reasoning tasks as an emergent property i.e., a series of intermediate reasoning steps lead the model to the final output known as \textit{chain-of-thought-prompting}. The parametric count of these models typically ranges from a few million to billions, the performance of which follows strict scaling laws \cite{b14}, with the larger models being more sample efficient and only mildly prone to overfitting.\\ 

\noindent The transformer architecture has achieved immense success in NLP, generative modeling \cite{b2, b10} and reinforcement learning tasks. The self-attention mechanism is the central idea of the transformer which help capture long-term dependencies, and can thus learn coherent patterns and capture the compositional relationships between them given as:
$$
Attention(Q, K, V) = softmax(QK^{T}/d_{k})V
$$

\noindent where $Q, K , V$ represents the input query, key and value matrix and $d_{k}$ denotes the key dimension respectively. Thus, they produce cohesive and plausible text when adequately tuned to the right temperature ($0 \leq T \leq 1$) during model inference and trained for several epochs ($>20$). This challenges the LMs ability to extrapolate and draw useful information from the training data to gain knowledge about the physical world to generate the correct answers. We believe that the statistical patterns implicitly present in the pre-trained text corpus is known to provide the model with a few innate priors and experience and is a possible explanation for its performance on few-shot learning methods such as the GPT-$3$ model \cite{b10, b13}.\\ 

\noindent In regard to commonsense and arithmetic reasoning tasks \cite{b24, b28, b29} large-language models (LLMs) such as GPT-3, Flacon, Llama achieve excellent performance on these tasks while fails to achieve good generalization as the model size is decreased. This leads to misrepresentation of knowledge and factual data in LMs. 

\subsection{Contribution}
\noindent In this regard, we evaluate the GPT-neo model in a relatively low-parameter regime, and subsequently, we gauge its performance in comparison to other baselines with larger model size. Our main contribution in this paper is as follows:\\

\begin{itemize}
    \item We investigate the generalization capabilities of the GPT-neo model through the commonsense reasoning lens. In particular, we employ a supervised learning objective discussion and test the model on a suite of $6$ tasks, namely Piqa, Winogrande, Hellaswag, Storycloze, BoolQ and OpenBookQA.\\
    \item We conduct adequate comparisons with competitive baselines such as GPT-$3$, Llama and Falcon for all our tasks in both zero-shot setting and fine-tuning methods.\\
    \item We also conduct extensive robustness tests and visualisations to assess its ability for \textit{in-context learning} and examine the overfitting concerns in Section $3$ \& $4$ respectively.\\
\end{itemize}

\noindent Our \textit{main goal} is not to demonstrate state-of-the-art results but to show that the GPT-neo model is competitive with its counterpart, such as the GPT-$3$ model, even though it is much smaller by a factor of $64x$.

\section{Can LLMs Reason?}
As discussed in \cite{b53, b54}, reasoning encompasses the ability of LMs for deduction, induction, abduction, analogy, commonsense and systematic methods for \textit{rational} problem solving in multi-steps of inference. This enables the model with \textit{abstraction} - allowing for the model to generalize to unseen test examples.\\

\noindent While LLMs are not trained explicitly to reason, it is more often an \textit{emergent} behaviour where the model mimics true reasoning abilities through less robust and generalizable mechanism - such as memorization or pattern matching using the contexts seen in the training data.\\

\noindent For example, \cite{b55} examines the in-context performance of GPT-$3$ on arithmetic tasks. They assess the model performance as a function of the frequency of the test samples in the training set. They observe that the model performs significantly lower for numbers not present in the training data and that thus lack the general ability to perform arithmetic. In addition, they demonstrate that superior performance on these tasks is rely largely on \textit{memorization} and mimics reasoning-like abilities by matching patterns present in the pre-training data.\\

\noindent Similarly, to test the memorization hypothesis, \cite{b56} construct "counterfactual tasks" to assess if LMs perform \textit{faithful} reasoning on commonsense and arithmetic tasks. Their results reveal degradation in performance on several counterfactual tasks and they attribute this gap to overfitting and absence of generalization, even though it requires the same reasoning ability as the original task.

\section{MODEL GENERALIZATION}
\noindent \cite{b3} defines generalization as the sensitivity to abstract analogies and thus requires a series of intermediate reasoning steps for complex downstream tasks. It involves adapting to novel situations by converting past experience into future skills. In this regard, neural networks are long known to perform well as function approximators on smooth (continuous) data manifolds without any discontinuities using gradient descent methods. Thus, this paper is an attempt in this direction to explore at least a few of these intuitive questions by resorting to the commonsense reasoning benchmark as a standard of measure in all our experiments.\\

\noindent It is believed that the success of LLMs is mainly due to \textit{in-context learning}: the ability to predict later tokens is easier because the auto-regressive model can now utilize the data present in-context i.e., the earlier tokens in the input data ($x_{i-1}$) itself \cite{b35} and \textit{induction heads}: mainly prefix matching (similar to skip-grams) and thus increases the logits of attended tokens \cite{b36}. This is known to drive meta-learning and the models ability to extract useful information from earlier context allowing information to move forward to attention heads in the subsequent layers.\\

\noindent In a recent work, \cite{b48} investigate the trajectory of the training process in masked language models (MLMs) to understand the phase transitions and specialised attention heads for emergent behaviour at different points in training. In particular, they introduce \textit{Syntactic Attention Structure} to examine the learning of complex linguistic phenomenon in MLMs which leads to model interpretability.\\

\noindent Likewise, \cite{b49} introduce distributional simplicity bias that identify the features of the training data that influence the network. They argue that neural networks learn higher order statistics and correlations only later in the training process which deviates from the general conception of model progression from linear to non-linear functions during stochastic gradient descent training.\\ 

\noindent \cite{b32} study in-context learning using selection algorithms of LLMs on various generation tasks for multiple input sequences without explicit prompting of the right task. This also allows for meta-learning and adapting the parameter initialization within few epochs of fine-tuning on downstream tasks.\\

\noindent It is observed that as the model size increases, the network can learn concepts that they pass on to downstream tasks and thus generalise to unseen instances than those seen during training through transfer learning as discussed in \cite{b10, b14, b22}. However, \cite{b19} postulate the model-wise double descent phenomenon where, as we increase the model size the performance first decreases i.e., a U-like behaviour and then increases on the test data. The former behaviour is expected because of the classical bias-variance tradeoff from statistical learning theory given as:
$$
MSE = bias^{2} + variance
$$

\noindent where $MSE$ is the mean-squared error. Thus, larger models exhibit lower bias with higher variance. Therefore, as the model size increases beyond a certain threshold called the over-parameterization regime, we expect the generalization error on the test dataset to increase while \cite{b19} argues that neural networks exhibit the opposite i.e., the generalization error decreases with increase in model complexity.\\

\noindent \cite{b23} examined the the models robustness with increasing capacity under projected-gradient descent adversarial examples. They observed that as beyond a certain point as we increase the size of the network there is a sharp transition in the model behaviour with steady increase in the training accuracy under strong adversaries.\\

\noindent Likewise, \cite{b20} observes that in the over-parameterization regime i.e., when the model complexity is large enough to cause zero training error leads to a decrease in the average test error and the performance of the worst group error i.e., the minority samples. The best worst-group error was observed for model with in the underparametrized regime with non-zero training error. This is attributed to the memorization of the minority samples in the training set and thus leads to poor generalization on unseen examples.\\

\noindent One way to test GPT-neo's ability in the fine-tuning phase with limited dataset constraints is through simple commonsense reasoning tasks, which requires recognizing a novel pattern that is unlikely to have occurred during pre-training and thus adapting quickly to the given task.\\ 

\noindent We demonstrate the effectiveness of our approach on a diverse array of tasks for natural language understanding. The results show that the model leverages the feature representations learned during unsupervised pre-training for larger performance gains during downstream tasks.\\

\begin{figure}[htp!]
\vspace{-0.7em}
\centering{\includegraphics[scale=0.2]{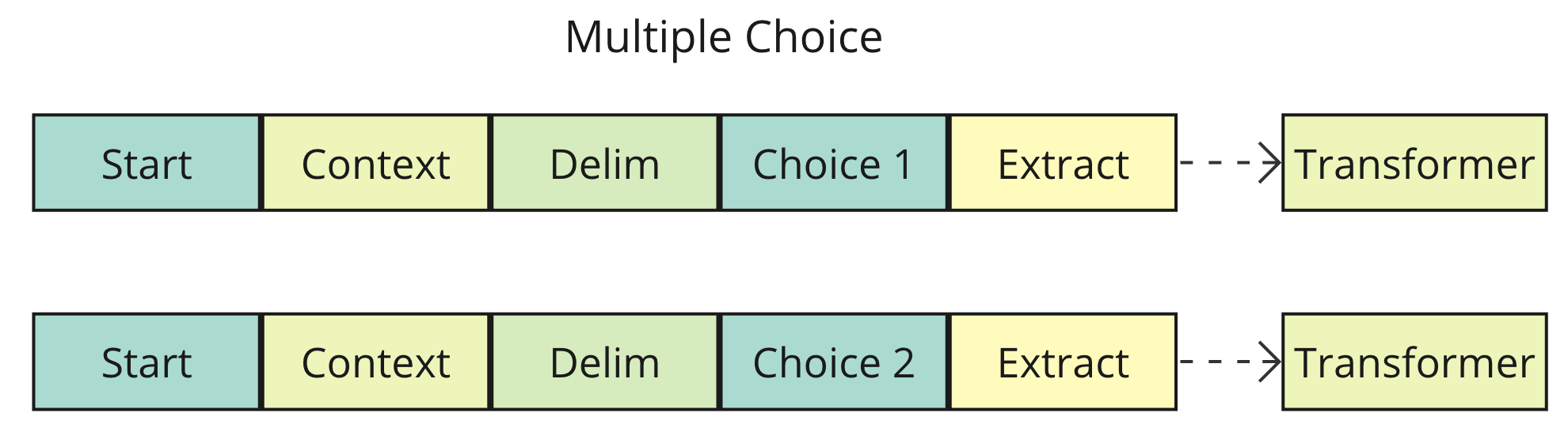}}
    \caption{Input format for GPT-neo architecture for multiple-choice task. Start and delim indicates the start and delimiter token respectively. }
    \label{fig:input}
    \vspace{-1em}
\end{figure}

\begin{table*}[htp!]
\resizebox{\linewidth}{!}{%
\centering
\begin{tabular}{l p{0.8\textwidth}}
\toprule[1.1pt] \\ [-2ex]
\textbf{Winogrande} & The trophy doesn't fit into the brown suitcase because the \_\_ is too large. \\
\medskip
& Choice 1. \textbf{Trophy} \quad Choice 2. Suitcase \\
\toprule[1.5pt]  \\ [-2ex]
\textbf{Piqa} & To make a tomato sauce taste more like a pizza sauce, \\
& Choice 1. \textbf{Mix extra paprika into the sauce to brighten the flavor.} \\
\medskip
& Choice 2. Mix a little bit of sugar into the sauce to sweeten it. \\
\toprule[1.5pt] \\ [-2ex]
\textbf{HellaSwag} & Making a cake: Several cake pops are shown on a display. A woman and girl are shown making \\
& Choice 1. \textbf{Bake them, then frost and decorate.} \\
& Choice 2. Taste them as they place them on plates. \\
& Choice 3. Put the frosting on the cake as they pan it. \\
\medskip
& Choice 4. Come out and begin decorating the cake as well. \\
\toprule[1.5pt] \\ [-2ex]
\textbf{StoryCloze} & I was walking to my house after riding the school bus. When I arrived at the front, I tried to open \\
& the door. The door was locked. When I tried to look for my keys, I couldn't find them anywhere. \\
& Choice 1. \textbf{I then found my spare key and entered.} \\
\medskip
& Choice 2. I was locked out and had to move to a new home. \\
\toprule[1.5pt] \\ [-2ex]
\textbf{BoolQ} & In Australia, each state has its own constitution. Each state constitution preceded the \\
& Constitution of Australia as constitutions of the then separate British colonies, but all the states \\
& ceded powers to the Parliament of Australia as part of federation in 1901. \\
& Question: Does each Australian state have its own constitution? \\
\medskip
& Answer: \textbf{True} \\
\toprule[1.5pt] \\ [-2ex]
\textbf{OpenBookQA} & Context: A stem is used to store water by some plants. \\
& Question: Stems are to flowers as \\
& Choice 1. Dogs are to cats. \\
& Choice 2. Cows are to cud. \\
& Choice 3. Bees are to pollen. \\
\medskip
& Choice 4. \textbf{Silos are to grains.} \\
\toprule[1.1pt] \\ [-2ex]
\end{tabular}}
\caption{An example is specified for each of the tasks used for evaluating the GPT-neo model with the correct answer choice indicated using the \textbf{bold} mark.}
\label{tab:example}
\end{table*}

\subsection{Related Work}
Recent work on evaluation of LLMs for commonsense natural language inference has garnered significant attention for larger models under zero-shot and few-shot settings. \cite{b40} conduct a comprehensive examination on four commonsense benchmark tasks. Similarly, \cite{b43} study LLMs under zero-shot settings and strong supervision, where they assess the model performance under various design choices such as input format and score function to better understand scaling laws of large models on NLI tasks.\\

\noindent \cite{b44} examine the proficiency of language models for transfer learning capabilities on downstream tasks using intermediate pre-training and adapter-based methods. Similarly, \cite{b45} introduces a weak supervision method for incorporating prompts into answer choices for effective knowledge transfer using a question-answering format i.e., using noisy predictions to get final predictions on commonsense benchmark tasks.\\

\noindent In contrast, our work is focused on assessing the performance of smaller language models against larger models and we also conduct numerous robustness tests and visualisation to investigate the discrepancy between these models as the size increases.\\

\noindent Our choice for evaluation of the GPT-neo model is motivated by the use of local and linear self-attention \cite{b37}, Mixture of Experts (MoE) \cite{b38} and axial positional embeddings \cite{b39} in the model architecture which aids in better \textit{in-context learning} and also lesser \textit{computational overhead} while training on TPUs.

\section{Dataset}
\label{dataset}
\noindent Commonsense tasks provide a way to understand the model's capability to interpret complex structural patterns that are not explicitly mentioned in the input text and are not part of the pre-trained data. The commonsense reasoning ability of large pre-trained models largely depends on the dataset quality and requires high levels of abstract reasoning capabilities to solve these tasks \cite{b34, b35}. These tasks include reading-comprehension, question-answering, sentence completion and classification. For multiple-choice tasks, we concatenate the context, question, and answer choice for each example to obtain $k$ different such sentences ($k$ choices) and pass it as input to the model as shown in Figure \ref{fig:input}. We compute each choice's probability and the softmax scores from their logit values.\\

\noindent For our experiments, we consider six diverse standard commonsense reasoning benchmark datasets (see Table \ref{tab:example}) and examine how well the model adapts to each of these tasks.\\ 


\noindent For example, in the sentence, \textit{"The corner table in a restaurant ordered a beer. The waiter served them the drink".} Humans can quickly establish that it is the people at the corner table that ordered the beer, and \underline{\textit{them}} refers to the people and not the corner table because we humans have a hard-coded notion of what people are. On the other hand, language models cannot quickly capture such implicit information or knowledge and hence fall back significantly on these tasks. Although relatively trivial for humans, these tasks are extremely hard for machines that merely rely on statistical patterns without proper understanding and abstraction capabilities. This paper shows that popular approaches to large-scale language pre-training, while highly successful on many abstract tasks, fall short when a physical world model is required.

\subsection{Winogrande}
Winogrande \cite{b4} is inspired by the "Winograd Schema Challenge" with increased task complexity. It consists of constructing two identical questions with two answer choices such that it includes a \textit{trigger word} which flips the corresponding answer questions. The task is to find a suitable entity for the pronoun. The training set consists of $44$k examples. Ex: \textit{"The Trophy doesn't fit into the brown suitcase because the \underline{$trophy^{*}/suitcase$} is too large".} Here, predicting what \textit{blank} requires the inference regarding the relative size of the trophy and the suitcase.


\subsection{Piqa}
PIQA \cite{b6} (Physical Interaction Question Answering) dataset is used to evaluate language representations and their knowledge about our physical world. This dataset describes how an object is built, used, or operated, which requires physical reasoning capabilities to select the right choice. It consists of over $16k$ training question-answering pairs, with an additional $2k$ and $3k$ validation and test examples respectively. As shown in Table \ref{tab:example}, the prediction of right choice as "\textit{mix extra \underline{paprika} into the sauce to brighten the flavor}", requires in-context learning and establishing coherent relations between \textit{tomato sauce, paprika and the pizza sauce}.

\subsection{StoryCloze}
Story Cloze \cite{b7} dataset involves selecting a plausible ending to a long story which is framed as a multiple-choice question consisting of $2$ answer choices. It consists of $4k$ examples, which are based on everyday events of daily life. It helps evaluate the extent to which the model learns causal and temporal relations between different entities within the given context.

\subsection{HellaSwag}
HellaSwag \cite{b8} is a commonsense natural language inference task with a data format similar to the Swag dataset but of higher quality. The hellaswag dataset involves picking the best \textit{ending} for a story. For each input question, the model is presented with the context from a caption and four choices to predict a what might happen next in a  coherent way. It contains a total of $50k$ sentences with an average length of $230$-word tokens.

\subsection{BoolQ}
    BoolQ is a question-answering dataset with yes/no answer choices containing $16k$ examples. Each example consists of a reading comprehension, question and an answer, with the title as additional context. The task entails various inferences to be drawn from the passage to find the correct answer choice, since the questions are generated in unprompted and unconstrained setting. In Table \ref{tab:example}, inferring the answer choice requires establishing adequate casual relations between the \textit{consitution of australia} and \textit{\underline{the federation act} in 1901.}

\subsection{OpennBookQA}
OpenBookQA is an advanced question-answering task used for commonsense NLI and consists of a $6k$ examples. It is modeled as an open-book exam where each example consists of a set of elementary science facts (context), a question and $4$ answer choices. As shown in Table \ref{tab:example}, predicting the right choice requires a deeper understanding of the input text comprehension and a general commonsense knowledge to logically connect the facts presented as context and establish the analogy as "\textit{\underline{silos are to grains}}".


\section{Model}
The GPT-neo model was released by Eleuther AI. The GPT-neo model is a transformer-based decoder-only autoregressive language model that uses a causal self-attention mechanism to learn contextual representations of individual word tokens. The GPT-neo was pre-trained using the standard language modeling objective: \textit{next token prediction} using causal self-attention. The objective is to maximize the log-likelihood of the data given as:
$$
L(\theta) = \sum_i \log p(x_{i} | x_{i-1}, ..., x_{2}, x{1} ; \theta)
$$
where $p$ the conditional probability of $x_{i}$ given its context \( \left\{ x_i\right\}_{i=1}^{i-1} \) which is modeled using the GPT-neo model with parameters $\theta$. The model is trained using gradient descent method.\\ 

\noindent The GPT-neo model is similar to the GPT-$3$ model but with key differences. It was pre-trained using on the Pile dataset \cite{b1} with consists of a diverse collection of $22$ high-quality datasets derived from numerous sources and with a similar performance boost compared to the GPT-$3$ models pre-trained on the common crawl dataset as the model size increases. This includes several datasets such as the books$3$, Pile-CC and DM-Mathematics and shows significant gains when fine-tuned on downstream tasks.\\

\begin{figure}[hbt!]
\centering{\includegraphics[scale=0.3]{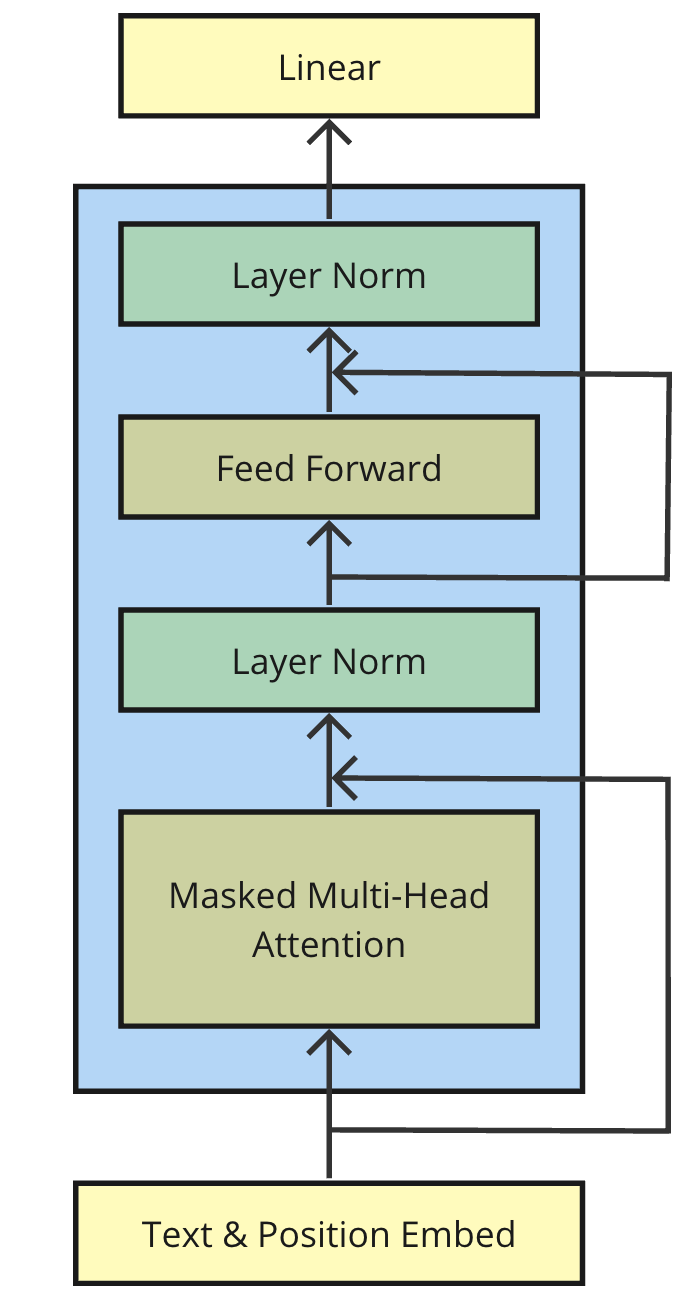}}
    \caption{GPT-neo architecture for multiple-choice task.}
    \label{fig:model}
    \vspace{-1em}
\end{figure}

\noindent We adapt the model parameters to the supervised learning task. Let  $\mathcal{C}= \left\{ x_i, y_i\right\}_{i=1}^m$ denote the set of $m$ training examples. The objective is to maximize the conditional probability $P(y|x)$ over the entire training set given as:
$$
l(\mathcal{C})=\sum_{(x, y)} \log P\left(y \mid x^1, \ldots, x^m\right)
$$
First, the input tokens are passed through the model to obtain the final transformer block's activation  $h_{l}$ and is then fed to a linear classifier to get model predictions $P_{i}$ given as:\\
\begin{equation*}
\begin{aligned}
h_0 & = A W_e + W_{ps} \\
h_l & =\operatorname{transformer} \_ \text {block }\left(h_{l-1}\right) \forall i \in[1, n] \\
P(i) & =\operatorname{softmax}\left(h_n W_e^T\right)
\end{aligned}
\end{equation*}

\noindent where $A$ is the context vector for tokens, $W_{e}$ is the embedding matrix and $W_{ps}$ is the positional-encoding matrix. We are primarily interested in evaluating: (1) the accuracy and (2) the cross-entropy loss on the test dataset given as:
$$
\mathcal{L} = - \sum_{i=1}^{d} y_{i} \log \hat{y_{i}} 
$$
where $y_{i}$ and $\hat{y_{i}}$ correspond to the true and predicted class labels respectively. We also examine the model robustness under adversarial attacks at sentence-level and visualize the attention maps in Section 3 and 4 respectively.\\



\section{Baselines}
\label{baselines}
We compare the performance of the GPT-neo model with several model baselines. The GPT-3 model is our primary choice since it offers similar choices in model sizes ($125$M, $350$M, $1.5$B \& $2.7$B) for comparison against the GPT-neo model. We evaluate the GPT-$3$ model under zero-shot and few-shot settings for all our tasks and utilize the standard hyperparameters as discussed in \cite{b10}.\\ 

\noindent We also assess the model performance against a family of pre-trained LLMs such as MPT ($7$B), Falcon ($7$B), Llama ($7$B \& $13$B) and GPT-J ($6$B) in both zero-shot setting and discriminative fine-tuning  for adequate model comparisons. We also utilize bidirectional language models such as BERT-base ($110$M) and BERT-large ($340$M) and report its accuracy scores for all our tasks as shown in Table \ref{tab:small models} $\&$ \ref{tab:large models}.

\begin{table*}[ht]
\resizebox{\linewidth}{!}{%
\centering
\renewcommand{\arraystretch}{1.3}
\begin{tabular}{|c|c|c|c|c|c|c|c|c|c|c|c|c|c|c|c|c|c|c|c|c|c|c|c|c|c|c|c|}
\hline
\backslashbox{Method}{M} & \multicolumn{1}{c|}{\textbf{MPT}} & \multicolumn{1}{c|}{\textbf{Falcon}} & \multicolumn{2}{c|}{\textbf{Llama-1}} & \multicolumn{2}{c|}{\textbf{Llama-2}} & \multicolumn{4}{c|}{\textbf{Fairseq}} & \multicolumn{5}{c|}{\textbf{GPT-3 Zero Shot}} & \multicolumn{5}{c|}{\textbf{GPT-3 Few Shot}} & \multicolumn{1}{c|}{\textbf{GPT-J}} & \multicolumn{1}{c|}{\textbf{BERT}} & \multicolumn{1}{c|}{\textbf{BERT-L}} & \multicolumn{4}{c|}{\textbf{GPT-Neo}} \\
\hline
& \textbf{7B} & \textbf{7B} & \textbf{7B} & \textbf{13B} & \textbf{7B} &\textbf{13B} & \textbf{125M} & \textbf{1.3B} & \textbf{2.7B} & \textbf{13B} &\textbf{125M}  &\textbf{350M} &\textbf{1.3B}  &\textbf{2.7B}  &\textbf{13B}  &\textbf{125M} &\textbf{350M} &\textbf{1.3B}  &\textbf{2.7B}  &\textbf{13B}  &\textbf{6B}  &\textbf{110M}  &\textbf{340M}  &\textbf{125M}  &\textbf{350M}  &\textbf{1.3B} &\textbf{2.7B} \\
\hline
\textbf{Piqa} & 79.3 & 75.4 & 78.1 & 80.0 & 78.0 & 80.3 & 66.1 & 72.3 & \textbf{75.0} & 75.6 & 63.5 & 69.1 & 74.1   & 74.6 & 78.4 & 62.3  & 68.1 & 73.2 & \ci{74.9} & 78.6 & 76.1 & 66.1 & 65.9  & 
62.0 & 64.0 & 69.1 & 71.1  \\
\hline
\textbf{BoolQ} & 74.9 & 66.5 & 74.5 & 77.1 & 76.9 & 80.6 & 55.9 & 57.6 & 60.2 & 64.1 & 48.6 & 59.8  & 61.4  & 66.5  & 65.7 & 42.9 & 59.4 & 63.7 & \ci{69.2} & 69.1 & 71.1 & \textbf{76.2} & 61.9 & 42.6 & 46.7 & 53.7 & 59.3\\
\hline
\textbf{Hellaswag} & 75.9 & 73.7 & 75.4 & 71.2 & 76.9 & 79.6 & 29.1 & 43.7 & 48.4 & 54.4 & 32.7 & 41.7 & 54.1 &  \textbf{62.1} & 69.0 & 32.4 & 42.7 & 53.4 & \ci{61.3} & 70.5 & 65.7 & 39.4 & 45.1 & 28.1 & 30.1 & 38.6 & 41.6 \\
\hline
\textbf{Winogrande} & 67.9 & 65.8 & 69.1 & 72.0 & 68.2 & 71.7 & 50.0 & 60.2 & 61.5 & 66.5 & 51.9 & 51.1 & 56.5 &  \ci{61.8} & 68.0 & 50.2 & 51.5 & 58.1 & 61.5 & 67 & 63.3 & 60.2 & \textbf{62.9}  & 50.0 & 51.1 & 54.0 & 55.5\\
\hline
\textbf{Storycloze} & 82.9 & 80.3 & 81.1 & - & 82.6  & - & 65.0 & 70.9 & 75.8 & 76.2 & 62.3 & 66.5 & 72.4 & 74.2 & 74.1 & 61.5 & 68.2 & 75.1 & \ci{79.2} & 82.3 & 55.3 & 65.4 & \textbf{80.5} & 45.6 & 51.0 & 54.7 & 59.2\\
\hline
\textbf{OpenBookQA} & 50.4 & 51.6 & 56.1 & 55.3 & 57.4 & 56.0 & 34.0 & 48.9 & 48.5 & 54.1 & 31.6 & 40.2 & 45.8  & 51.0  & 54.9 & 36.0 & 42.6 & 49.5 & \ci{54.2} & 59.8 & 50.2 & 52.1 & \textbf{61.4} & 35.5 & 39.2 & 42.2 & 46.3 \\
\hline
\end{tabular}}
\caption{Test accuracy for commonsense reasoning tasks using the GPT-neo model and model baselines. The best performer across all methods is denoted using the \textbf{bold} mark (excluding model size $\geq 6$B parameters). For ease of comparison, we color the second best performer with {\color{blue} blue} color.}
\label{tab:small models}
\end{table*}

\section{Experimental Setting}
\label{experimental setting}
\noindent  In this section, we give a general overview of the training procedure used for fine-tuning both GPT-neo and the baseline models. We perform all our model fine-tuning using Google Cloud TPU- VMs using the TPU-V8 accelerators. We use the JAX framework for model training and evaluation and obtain significant training speed boost-up using \textit{vmaps} and \textit{pmaps} for vectorisation and parallelisation on multiple devices. We train the models for $7$ epochs and use a fixed random seed throughout our experiments. We also add a dropout of $0.2$ for regularization. We use the byte pair encoding (BPE) tokenization scheme with a vocab size of $50257$ tokens.\\

\noindent For all our tasks, we use the adafactor optimizer with a learning rate of $2e-5$ to $3e-7$ and do not use gradient clipping. We use a linear learning rate decay schedule with warm-up over $0.1\%$ of training, and the value annealed $3e-7$. We slightly deviate from the standard approach used in GPT-$3$, where we use the maximum length of the input tokens for padding the input sequences. This does not affect the model performance and also results in a reduced memory footprint on the TPU accelerators.\\

\noindent We use a batch size ($k$) of $32$ i.e., $4$ examples on each device. Also through experiments on various batch sizes ranging from $k = [8, 16, 32]$ and conclude that the $32$ version outperforms significantly when experimented for each of our datasets.\\

\begin{table}[ht]
\centering
\begin{tabular}{|c|l|}
Parameter & Value\\
\hline
$n_{heads}$ & $16$\\
$n_{layers}$ & $24$\\
$d_{model}$ & $768$\\
$n_{linear}$ & $2048$\\
$d_{vocab}$ & $50257$\\
\end{tabular}
\caption{Training Parameters used for GPT-neo model training.}
\label{tab:equations}
\vspace{-1em}
\end{table}

\noindent In this section, we assess GPT-neo's performance on a suite of commonsense reasoning benchmark tasks that involve sentence completion, mulitpl-choice natural language inference, using the setting discussed in Section \ref{experimental setting}. We also report the model scores for baselines (Section \ref{baselines}) to ensure comparisons across numerous model sizes.\\

\begin{figure}[htp!]
    \vspace{-0.6em}
    \centering
    \includegraphics[width=0.6\columnwidth]{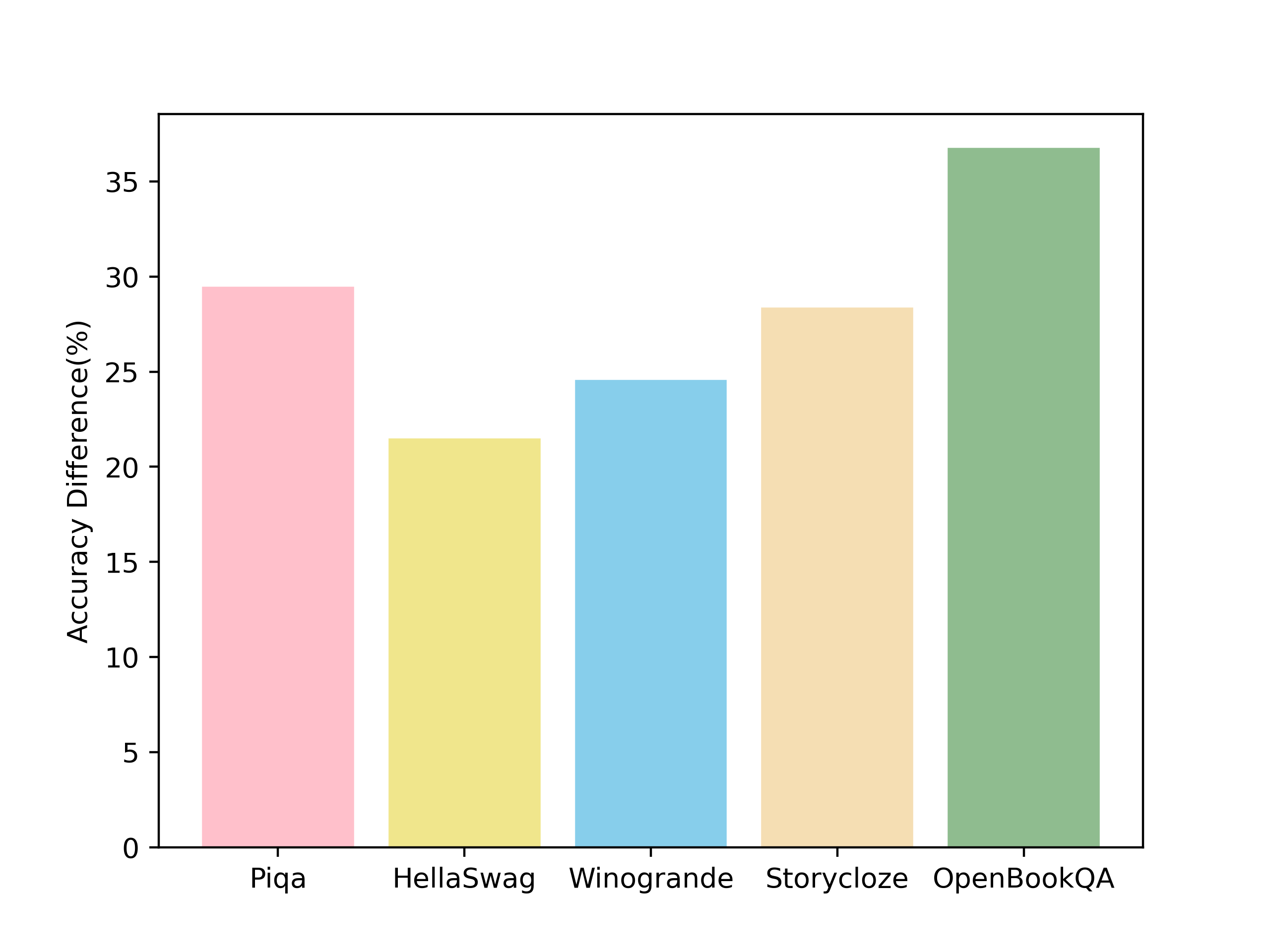}
    \caption{Performance comparison of GPT-neo model and Random baselines for each of our task.}
    \label{random baselines}
\end{figure}

\noindent Figure \ref{random baselines} compares the performance gap of the GPT-neo model against random baselines for all our tasks. The random baseline is the GPT-neo model without discriminative fine-tuning where the probability of choosing the correct answer choice is given as $p(y|x) = 1/k$ where $k$ is the number of choices.


\section{Results}

\noindent In the following section, we present our results across for all the tasks discussed in Section \ref{dataset}. Please note that we exclude the results for the $13$B parameter model in the following discussion, as our largest model is of size $2.7$B parameters. However, the results for the $13$B parameter model are presented in Table \ref{tab:small models} for reference.

\subsection{Winogrande}
We evaluate our results on the Winogrande test dataset and obtain an accuracy of $55.5\%$ which  is competitive with the GPT-$3$ model. We also observe that the model accuracy shows improved performance with an $5.5\%$ increase in accuracy as the size increases from $125$M to $2.7$B model parameters. However, we observe that the GPT-$3$ few-shot and BERT-large model achieve the Top-$2$ accuracy of $61.8\%$ and $62.9\%$ respectively (excluding models $\geq 6$B), which is attributed to their large model size.

\subsection{Piqa}
Here, we achieve an accuracy of $71.1\%$ upon fine-tuning, which is only slightly lower than the GPT-3 few-shot and the Fairseq model by $4\%$. We observe that the model is competitive with both the Fairseq and GPT-3 model as the model size is increased and achieves a performance drop of $4\%$ and $8\%$ with respect to the Falcon and the MPT model respectively.

\subsection{HellaSwag}
We obtain an accuracy of $41.6\%$ on the Hellaswag dataset which is our worst performance among all the tasks. The GPT-$3$ model significantly outperform our model by $19\%$ in both the zero-shot and few-shot setting. For model size of $13$B, Llama-$1$ and Llama-$2$ obtain excellent results with a model best accuracy of $79.6\%$ which is $9\%$ higher than the GPT-$3$ few-shot results.

\subsection{StoryCloze}
On the StoryCloze dataset, our model achieves an accuracy of $59.2\%$. In contrast to HellaSwag, the GPT-$3$ model outperforms the GPT-neo model by a margin of $15 - 18 \%$, even when considering both the $125$M and $350$M versions.  Surprisingly enough the Fairseq model demonstrates better performance compared to the GPT-$3$ few-shot model for model sizes of $125$M and $350$M. However, GPT-$3$ excels in performance as its model size scales beyond $1$B parameters, achieving the highest accuracy of $82.3\%$ in the few-shot setting.

\subsection{BoolQ}
On this task, GPT-neo achieves an accuracy rate of $59.3\%$, while Llama-$2$ attains a significantly higher accuracy rate of $80.6\%$. While the model's performance remains competitive compared to the GPT-$3$ few-shot model with $125$M parameters, exhibiting only a marginal difference of $0.3\%$, it becomes evident that as model size increases, GPT-$3$ rapidly adapts to the task, showcasing a substantial $17.5\%$ performance improvement when transitioning from the $125$M to the $350$M model. In contrast, GPT-neo demonstrates a comparatively modest increase of $4.5\%$ over the same transition.

\subsection{OpenBookQA}
We obtain an accuracy of $46.3\%$ on our largest model which is comparable with the Fairseq model but the GPT-$3$ model is still better by $8\%$ in the few-shot setting. We also observe that BERT-large which contains just $340$M parameters obtains the highest accuracy of $61.4\%$ among all the models. Larger models such as MPT and Falcon obtain accuracy of $50.4\%$ and $51.6\%$ which is higher than the GPT-neo model by only $4\%$. The attention head visualization for the example presented in Table \ref{tab:example} is discussed in Section \ref{Visualization}.\\ 

\noindent We also report the mean accuracy scores for all the models reported in Table \ref{tab:large models} with model size  greater size $20$B parameters. Since, these models are computational expensive  and results in memory-overhead, we do not perform these supervised fine-tuning for models $\geq 20$B parameters, but report their scores from the literature \cite{b10, b26, b27, b42, b45}. We also include our original GPT-neo results from Table \ref{tab:small models} for comparisons with larger models. As expected the performance exhibits a significant and consistent improvement as the model size increases across all benchmark tasks.
\section{Robustness Test}
\begin{figure}[!tbp]
  \centering
  \begin{minipage}[b]{0.4\textwidth}
    \includegraphics[width=\textwidth]{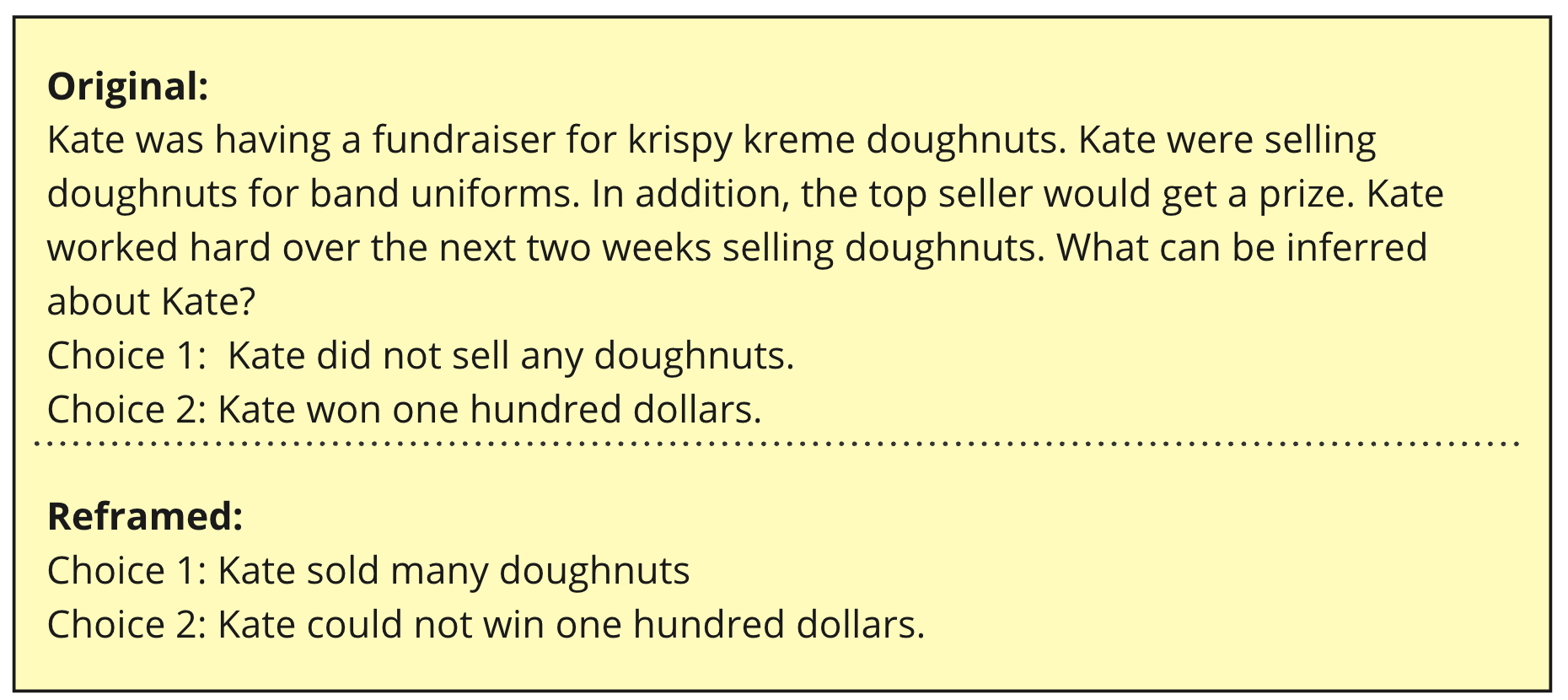}
  \end{minipage}
  \hfill
  \begin{minipage}[b]{0.4\textwidth}
    \includegraphics[width=\textwidth]{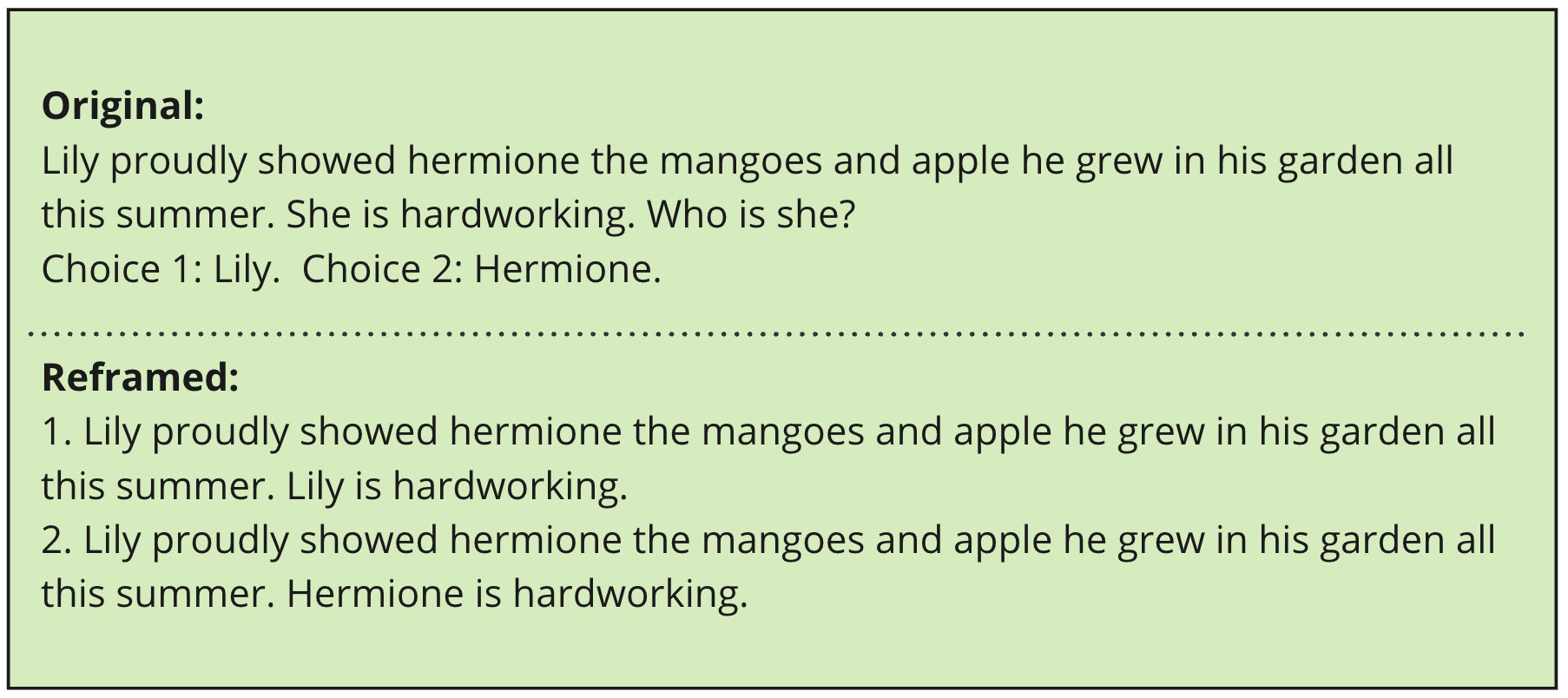}
    \caption{Reframed sentences-we construct additional samples by modifying the original sentence or its answer choice, which is used to test the model during inference time.}
  \end{minipage}
\end{figure}
\cite{b22} examine a universal law of robustness where they state that to truly memorize the dataset (in the sense of low training error i.e., $L_{train} << \epsilon$, where $\epsilon \rightarrow 0$) and to do so robustly (measured using the Lipschitz \footnote{A function $f:R^{d} \rightarrow R$ is $L$-Lipschitz with respect to the norm $\|.\|$ if $ \forall x,y \in R^{d}, |f(x) - f(y)| / \| x - y\| \leq L$. } constant $L$), requires that:
$$
p \geq nd
$$

\noindent where $p$ is the number of model parameters, $n$ is the number of high-dimensional data points and $d$ is the ambient input dimension respectively. This is called \textit{dramatic overparametrization}.\\

\noindent We are mainly interested in evaluating if our results are robust to the possibility that one of our assumptions might not be accurate. We perform the robustness test on a few examples to examine GPT-neo ability to  learn the right semantics at token-level.\\

\noindent We follow the conventional dual text instance method as discussed in \cite{b12, b42} by constructing a distorted sentence using the original test sample and thus inferring the robustness measure.Note that we perform the robustness tests for both these examples \textit{in-context} i.e., predictions at the sentence-level. Each task involves the addition of noise into the original sentences by means of addition, deletion and replacing of adequate word tokens and then predict the similarity scores between both these sentences with respect to the input context.\\ 

\begin{center}
\begin{table}[htbp]
\centering
\renewcommand{\arraystretch}{1.1}
\resizebox{0.45\textwidth}{!}{
\begin{tabular}{|l|c|c|c|c|c | l | ccc|}
\hline
\backslashbox{Method}{M} & \textbf{Piqa} & \textbf{BoolQ} & \textbf{HellaSwag} & \textbf{Storycloze} & \textbf{Winogrande} \\
\hline
\textbf{Addition} & 90  & 81 & 81 & \textbf{94} & 74\\
\hline
\textbf{Subtraction} & 80  & 62 & 63 & \textbf{85} & 62\\
\hline
\textbf{Replace} & 92  & 71 & 84 & \textbf{93} & 75\\
\hline
\textbf{Swap} & \textbf{96}  & 67 & 66 & 91 & 67\\
\hline
\end{tabular}}
\vspace{0.2em}
\caption{Estimation accuracy (in \%) for robustness tests across all the tasks.}
\label{tab:Estimation Accuracy}
\end{table}
\vspace{-1.5em}
\end{center}

\begin{table}[ht]
\resizebox{\linewidth}{!}{%
\centering
\renewcommand{\arraystretch}{1.3}
\begin{tabular}{|c|c|c|c|c|c|c|c|c|c|}
\hline
\backslashbox{Dataset}{M} & \multicolumn{1}{c|}{\textbf{MPT}} & \multicolumn{1}{c|}{\textbf{Falcon}} & \multicolumn{2}{c|}{\textbf{Llama-1}} & \multicolumn{1}{c|}{\textbf{Llama-2}} & \multicolumn{1}{c|}{\textbf{GPT-3}} & \multicolumn{1}{c|}{\textbf{GPT-Neo X}} & \multicolumn{1}{c|}{\textbf{GPT-Neo}} & \multicolumn{1}{c|}{\textbf{Human}} \\
\hline
& \textbf{30B} & \textbf{40B} & \textbf{33B} & \textbf{65B} & \textbf{70B} & \textbf{175B} & \textbf{20B} & \textbf{2.7B} &  \\
\hline
\textbf{Piqa} & 81.9 & 82.4 & 82.3 & \textbf{82.8} & \textbf{82.8} & 82.3 & 78 & 72.14 & 94.09  \\
\hline 
\textbf{BoolQ} & 79.0 & 83.1 & 83.1 & \textbf{85.3} & 85.0 & 77.5 & - & 57.3 & 90.0  \\
\hline
\textbf{Hellaswag} & 79.9 & 83.6 & 82.8 & 84.2 & \textbf{85.3} & 79.3 & 71.2 & 42.73 & 95.6 \\
\hline
\textbf{Winogrande} & 71 & 76.9 & 76 & 77 & \textbf{80.2} & 77.7 & 66.5 & 56.5 & 94.1  \\
\hline 
\textbf{OpenBookQA} & 52.0 & 56.6 & 58.6 & 60.2 & 60.2 & \textbf{65.4} & 32.6 & 56.5 & 92.0  \\
\hline 
\textbf{Storycloze} & - & - & - & - & - & \textbf{87.7} & - & - & 91.5  \\
\hline
\end{tabular}}
\vspace{0.2em}
\caption{Performance comparison of Large language models ($\geq 20$B) from the literature. We follow similar
convention as that in Table \ref{tab:small models}.}
\label{tab:large models}
\end{table}

\noindent For each of the robustness task, we create $100$ test samples using random sampling from the test dataset and forming a pair of coherent test sentences, testing for consistent results (\textit{defined} as $\%$ of test samples with accurate predictions on both the original and distorted sentences) across all the instances, and report their mean accuracy scores in Table \ref{tab:Estimation Accuracy}.\\ 

\noindent The goal of this task is to ascertain accurate predictions of the answer choice that corresponds to the original sentence when the distorted sentence maintains logical coherence, while allowing a random selection otherwise such as the \textit{subtraction} task. We test the model's capabilities on four tasks given as follows:\\

\begin{enumerate}
\item \textbf{Addition}: Addition of words to its sentences, such that they are semantically equivalent, and is required to output the same answer choice for both the sentences.\\
\item \textbf{Subtraction}: The polarity of the sentence is reversed to negate its semantic meaning, wherein the sentences contradict each other while the answers choices for both the original and new sentence remains the same.\\
\item \textbf{Swap}: The role of each noun swapped with one another and is expected to pick the right choices by contextual inference by identifying the roles of each noun respectively.\\
\item \textbf{Replace}: We replace its word tokens so that they are logically coherent and should output the right choice depending on the context presented in the sentence.\\
\end{enumerate}

\begin{figure}[htpb]
    \vspace{-0.6em}
    \centering
    \includegraphics[width=\linewidth]{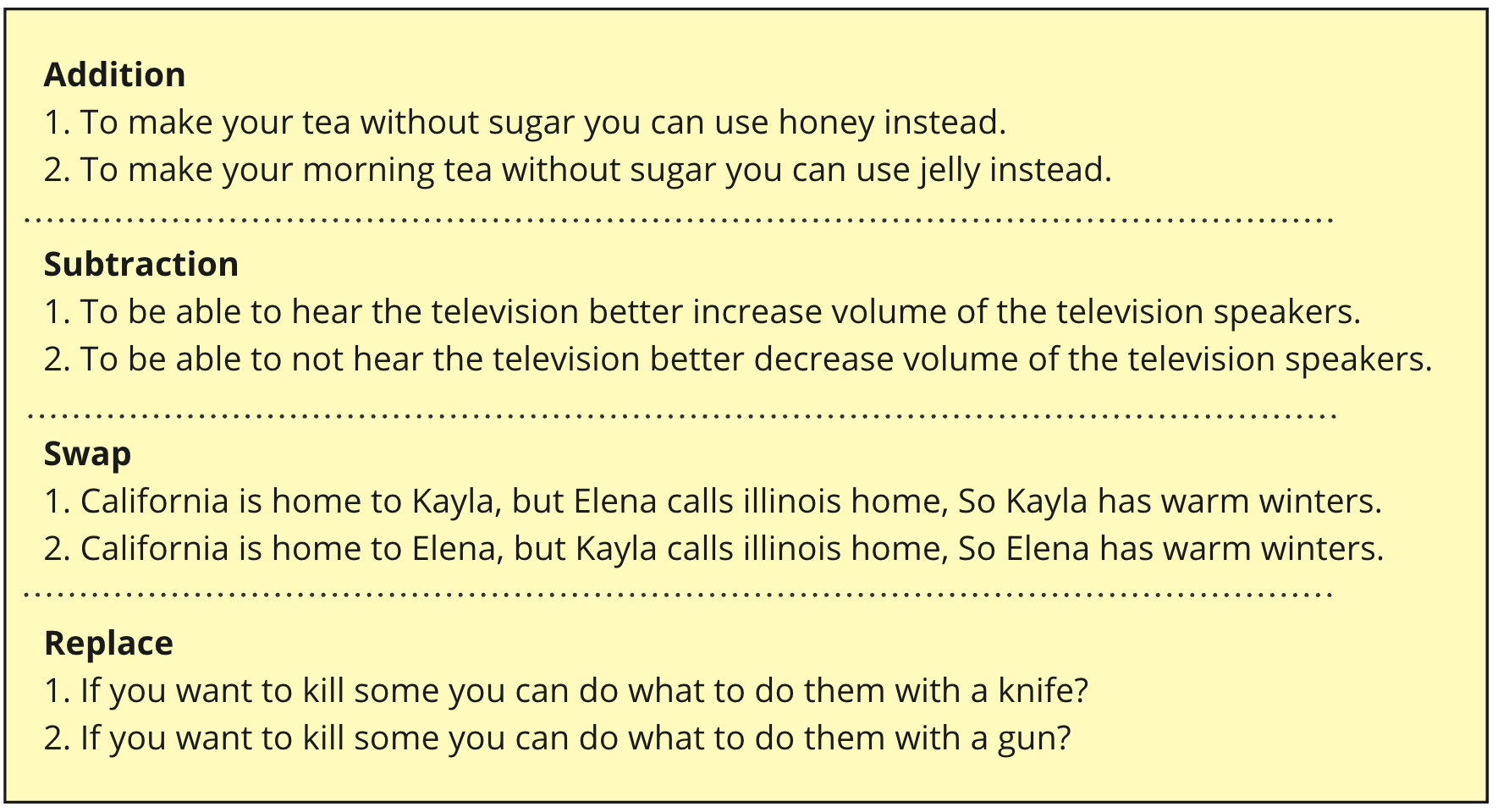}
    \caption{Samples used for measuring the robustness of the GPT-neo models using various methods such as addition, subtraction, swapping, and replacing. We use two sentences here to indicate the original (first row) and distorted (second row) samples, and the model accurately predicts the positive sentence and as correct or incorrect predictions for the negative sentence.}
    \label{Figure 2}
    \vspace{-0.1cm}
\end{figure}

\noindent In Table \ref{tab:Estimation Accuracy}, we present the mean estimation accuracy of the robustness test for various datasets. The reported accuracies correspond to the predictions obtained using the GPT-neo model in a zero-shot setting without further fine-tuning on the additional distorted test samples. The Piqa and Storycloze dataset performs consistently well when tested under different settings on all our tasks with accurate predictions. However, Hellaswag demonstrated strong performance in two specific tasks i.e., add and replace while exhibits a considerable drop in accuracy for the remaining tasks.\\

\noindent Likewise the performance on Boolq and Winograde dataset is somewhat degraded, with some overfitting concerns and its inability to adapt to the given context. A closer inspection of the incorrect responses reveals that the model frequently struggles to infer contextual meanings and at times is insensitive to such changes for both positive and negative sentences. Overall, GPT-neo displays reasonable proficiency when probed at various zero-shot settings for atleast a few of these tasks. \footnote{We do not report robustness test scores on the OpenBookQA dataset because plausible modifications to the original sentence was not possible while preserving the semantics with respect to the answer choices.}

\section{Visualization}
\label{Visualization}
\begin{figure}[htp!]
    \vspace{-0.6em}
    \centering
    \includegraphics[width=0.8\columnwidth]{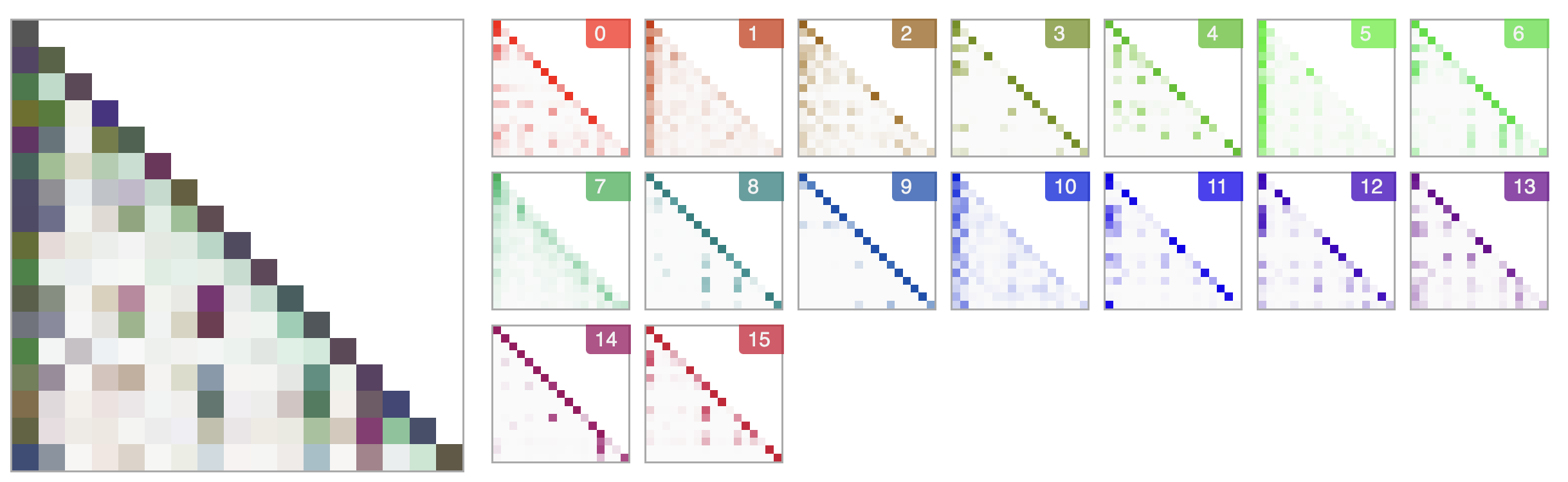}
    \label{attention heads}
\end{figure}

\begin{figure}[htp!]
    \vspace{-0.6em}
    \centering
    \includegraphics[width=0.8\columnwidth]{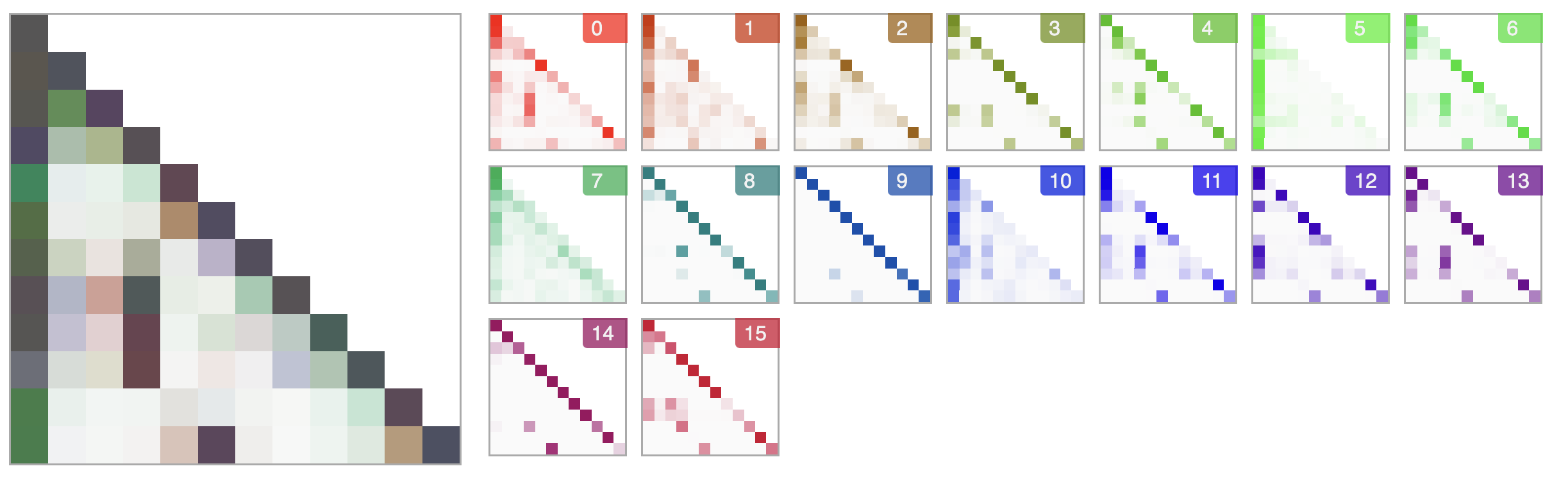}
    \caption{Attention head for GPT-neo model Layer $6$ for the example in the HellsSwag dataset as discussed in Table \ref{tab:example}.}
    \label{attention 1}
\end{figure}

\begin{figure}[htp!]
    \vspace{-0.6em}
    \centering
    \includegraphics[width=0.8\columnwidth]{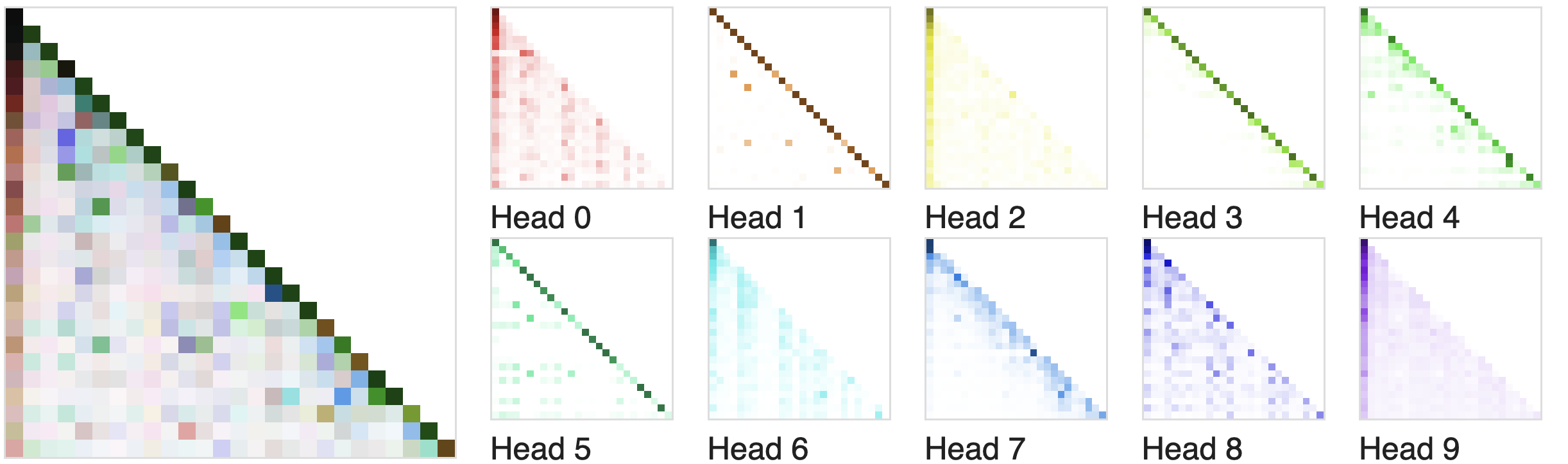}
    \caption{Attention head for GPT-neo model Layer $6$ for the example in the OpenBookQA dataset as discussed in Table \ref{tab:example}.} 
    \label{attention 2}
\end{figure}

\noindent In this section, we visualise the attention pattern using the GPT-neo model for sentence completion tasks using the source and destination token. We visualise the overall attention pattern and individual attention heads for a few examples as shown in Figure \ref{attention 1} $\&$ \ref{attention 2}. We are mainly interested in understanding \textit{in-context learning} and \textit{induction heads} i.e., a composition of attention heads that work together to copy patterns. It is observed that the lower layers of transformer models capture global token dependencies across multiple tokens  which enables the transfer of relevant information across layers and thus predict the correct answer choice \cite{b50, b51}.

\begin{figure}[ht]
    \vspace{-0.6em}
    \centering
    \includegraphics[width=0.8\columnwidth]{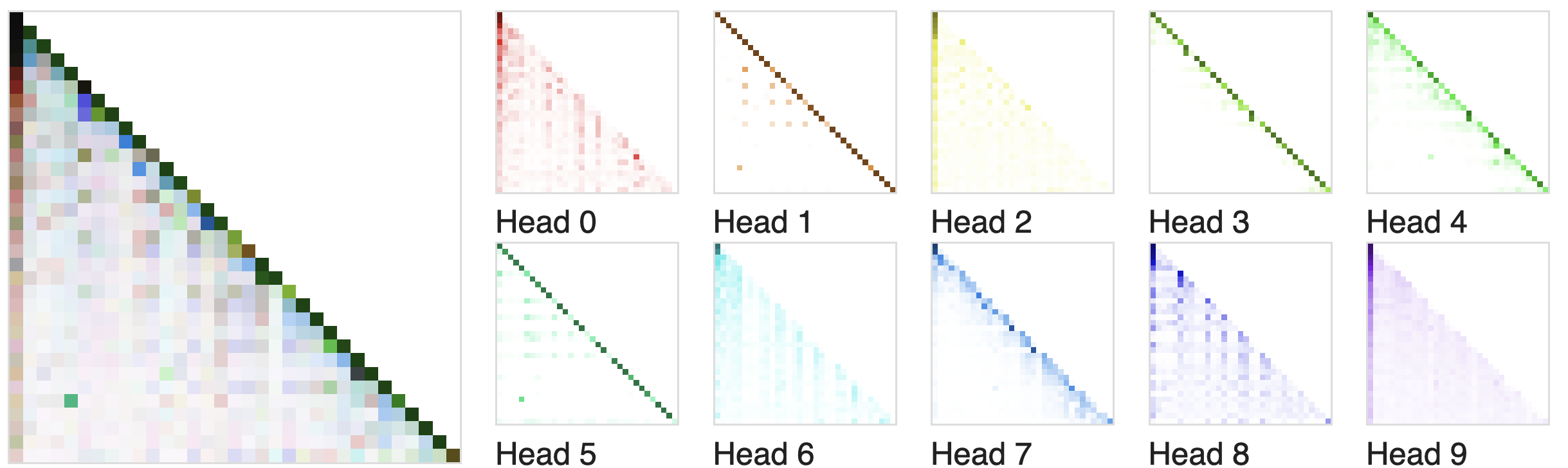}
    \label{Figure 2}
\end{figure}

\begin{figure}[htp!]
    \vspace{-0.6em}
    \centering
    \includegraphics[width=0.2\columnwidth]{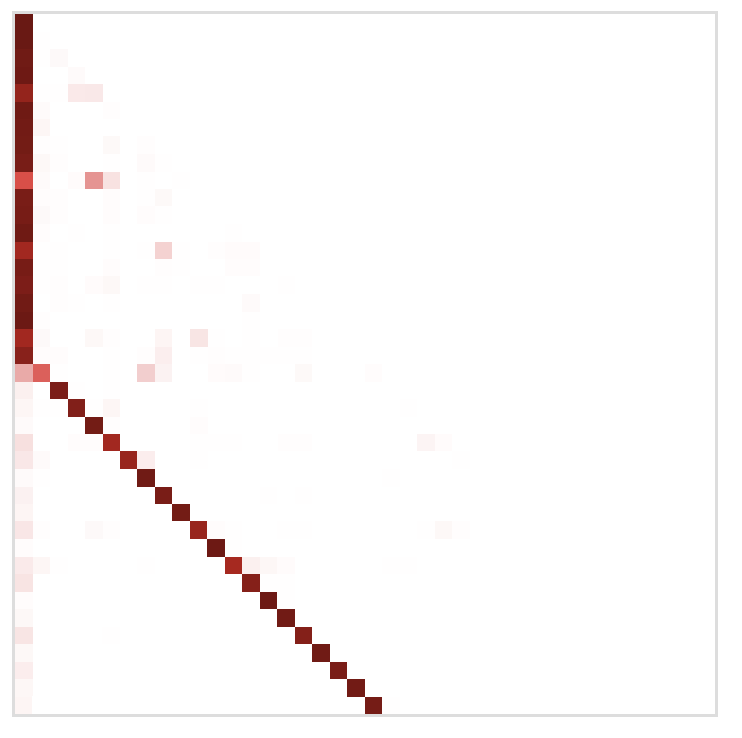}
    \caption{Attention hook for GPT-based LLMs as demonstrated in \cite{b52}.}
    \label{Attention hook}
\end{figure}

\noindent In Figure \ref{attention 1} $\&$ \ref{attention 2}, we observe \textit{previous token} attention heads that simply attend back to the preceding token and also \textit{first token} heads that fall back to the first token and do nothing as discussed in \cite{b50}. In comparison, we also find specialized attention heads i.e., the pattern is diagonal while most heads have no clear identifiable structure. In Figure \ref{Attention hook}, we demonstrate an interesting hook in the attention pattern activation for the relevant head as discussed in \cite{b52}.

\section{Limitation}
Although, we conduct the examination of the GPT-neo model, we understand that we comparisons are not exhaustive in terms of datasets considered and the ablation studies for understanding the model performance using robustness tests. Likewise, due to limited computational resource availability, we were unable to perform extended model training for for more epochs and supervised fine-tuning on large models in particular for the Storycloze dataset. In our subsequent work, we hope to extend this further to probe smaller models for a larger class of datasets with interpretable results. We hope our work provides some preliminary motivation and understanding on the impact of smaller models in the purview of larger models. 

\section{Conclusion}
In this work, we illustrate the commonsense reasoning capabilities of the GPT-neo model on a suite of $6$ diverse tasks. Furthermore, we incorporated various model baselines for comparative analysis and have demonstrated that the GPT-neo model delivers competitive performance on several tasks as the model size is increased. We also investigated the model using attention head visualization and conducted robustness tests to better understand the model performance under various settings.

\section{Acknowledgments}
We thank the anonymous reviewers for their review and suggestions.
\bibliographystyle{ACM-Reference-Format}

\vspace{12pt}

\end{document}